\documentclass[preprint,12pt]{elsarticle}

\usepackage{amsmath,amsfonts}
\usepackage{algorithmic}
\usepackage{array}
\usepackage[tight,footnotesize]{subfigure}
\usepackage{textcomp}
\usepackage{stfloats}
\usepackage{url}
\usepackage{verbatim}
\usepackage{graphicx}
\def\BibTeX{{\rm B\kern-.05em{\sc i\kern-.025em b}\kern-.08em
    T\kern-.1667em\lower.7ex\hbox{E}\kern-.125emX}}
\usepackage{balance}
\usepackage{booktabs}
\usepackage{caption}
\usepackage{makecell}
\usepackage{adjustbox}
\usepackage{wrapfig}
\usepackage[linesnumbered,ruled]{algorithm2e}
\usepackage{amssymb}
\usepackage{lineno}
\usepackage{float}
\usepackage{algorithmic}
\usepackage[tight,footnotesize]{subfigure}
\usepackage{booktabs}
\usepackage{caption}
\usepackage{makecell}
\usepackage{adjustbox}
\usepackage{wrapfig}
\usepackage{xcolor}
\usepackage[linesnumbered,ruled]{algorithm2e}
\usepackage{listings}
\usepackage{makecell}
\journal{Journal Name}

\begin{document}
\sloppy
\setlength{\parskip}{0pt}

\begin{frontmatter}




\title{AI Trust OS --- A Continuous Governance Framework for Autonomous AI Observability and Zero-Trust Compliance in Enterprise Environments}


\author[label1]{Eranga Bandara}
\ead{cmedawer@odu.edu}

\author[label10]{Asanga Gunaratna}
\ead{founders@complianceoslab.app}

\author[label1]{Ross Gore}
\ead{rgore@odu.edu}

\author[label3]{Abdul Rahman}
\ead{abdulrahman@deloitte.com}

\author[label1]{Ravi Mukkamala}
\ead{mukka@odu.edu}

\author[label1]{Sachin Shetty}
\ead{sshetty@odu.edu}

\author[label7]{Sachini Rajapakse}
\ead{sachini.rajapakse@iciclelabs.ai}

\author[label7]{Isurunima Kularathna}
\ead{isurunima.kularathna@iciclelabs.ai}

\author[label1]{Peter Foytik}
\ead{pfoytik@odu.edu}

\author[label1]{Safdar H. Bouk}
\ead{sbouk@odu.edu}

\author[label4]{Xueping Liang}
\ead{xuliang@fiu.edu}

\author[label8]{Amin Hass}
\ead{amin.hassanzadeh@accenture.com}

\author[label5]{Ng Wee Keong}
\ead{awkng@ntu.edu.sg}

\author[label6]{Kasun De Zoysa}
\ead{kasun@ucsc.cmb.ac.lk}


\address[label1]{Old Dominion University, Norfolk, VA, USA}
\address[label10]{AI Motion Labs, Melbourne, Australia}
\address[label3]{Deloitte \& Touche LLP, USA}
\address[label4]{Florida International University, USA}
\address[label5]{Nanyang Technological University, Singapore}
\address[label6]{University of Colombo, Sri Lanka}
\address[label7]{IcicleLabs.AI}
\address[label8]{Accenture Technology Labs, Arlington, VA, USA}

\begin{abstract}
The accelerating adoption of large language models, retrieval-augmented generation 
pipelines, and multi-agent AI workflows within enterprise environments has created a 
structural governance crisis. Organizations deploying AI systems at scale face a 
fundamental accountability problem: they cannot govern what they cannot see, and existing 
compliance methodologies — built for deterministic, stateless web applications — provide 
no mechanism for discovering, classifying, or continuously validating AI systems that 
emerge organically across engineering teams without formal oversight. The result is a 
widening trust gap between what enterprise buyers, regulators, and boards demand as proof 
of AI governance maturity and what organizations can actually demonstrate. This paper proposes AI Trust OS, a conceptual framework and governance architecture for 
continuous, autonomous AI observability and zero-trust compliance in enterprise 
environments. Rather than treating AI governance as a periodic audit exercise driven by 
self-attestation and manual evidence collection, AI Trust OS reconceptualizes compliance 
as an always-on, telemetry-driven operating layer — one in which AI systems are discovered through observability signals, control assertions are collected by automated probes, and trust artifacts are synthesized continuously rather than assembled reactively under audit pressure. The framework is grounded in four foundational principles: proactive discovery over reactive declaration, telemetry evidence over manual attestation, continuous posture over point-in-time audit, and architecture-backed proof over policy-document trust. Architecturally, AI Trust OS operates through a zero-trust telemetry boundary in which ephemeral, read-only credential probes validate structural configuration metadata without ingressing source code, prompt content, or payload-level personally identifiable information. An AI Observability Extractor Agent continuously scans LangSmith and Datadog LLM telemetry streams, automatically registering undocumented AI systems and shifting the epistemological basis of enterprise AI governance from organizational self-report to empirical machine observation. The framework is evaluated through its governance coverage, discovery accuracy, evidence integrity, and compliance posture outcomes, and is situated within the evolving regulatory landscape of ISO 42001, the EU AI Act, SOC 2, GDPR, and HIPAA. The paper argues that continuous, telemetry-first AI governance represents not an incremental improvement over existing compliance tooling but a categorical architectural shift in how enterprise trust is produced, maintained, and demonstrated in the AI era.
\end{abstract}


\end{frontmatter}

\section{Introduction}

The enterprise software landscape is undergoing a fundamental transformation. Where 
organizations once deployed predictable, rule-based software systems whose behavior 
could be formally verified and statically audited, they now operate ecosystems of 
probabilistic, model-driven AI systems whose behavior is emergent, context-dependent, 
and distributed across multiple third-party vendors, inference endpoints, and data 
processing chains~\cite{sculley2015hidden}. Large language models generate outputs 
that cannot be fully anticipated from their inputs~\cite{bommasani2021opportunities, llm}. Retrieval-augmented generation pipelines traverse multiple vendor boundaries before 
producing a response~\cite{lewis2020retrieval}. Multi-agent workflows execute sequences 
of autonomous decisions that compound in ways no single policy document can fully 
describe~\cite{sapkota2025ai, agentic-ai, agentic-workflow-practicle-guide}.

This transformation has not been matched by a corresponding evolution in governance 
infrastructure. The compliance and security frameworks that enterprise organizations 
depend upon — SOC 2, ISO 27001, HIPAA, GDPR — were designed for a world of static 
web applications with deterministic data flows and auditable configuration 
states~\cite{disterer2013iso}. Their evidence collection methodologies assume that a human practitioner can manually gather screenshots of cloud consoles, compile those 
artifacts into workpapers, and present them to an auditor as proof of control 
effectiveness~\cite{raji2020closing}. That assumption breaks down entirely when the 
systems being governed are AI pipelines that span five vendors, process sensitive data 
through embedding models, retrieve context from vector databases, generate outputs through foundation models, and log everything to observability platforms — all within 
a single user request~\cite{weidinger2021ethical}.

The resultant governance deficit is not solely operational in nature; it is 
simultaneously commercial and regulatory. Enterprise procurement functions 
increasingly require real-time, empirical demonstration of AI governance maturity 
as a prerequisite for completing vendor assessments~\cite{mokander2023auditing}. 
Regulatory instruments, most notably the EU AI Act, have introduced legally binding 
risk-classification obligations for AI systems~\cite{euaiact2024}. In parallel, 
ISO 42001 has codified an AI management system standard that mandates organizations 
to maintain structured inventories of AI systems, conduct formally documented risk 
assessments, and provide demonstrable evidence of control implementation and 
coverage~\cite{iso42001}. Entities that are unable to furnish such evidence — or 
that can do so only ex post, under audit pressure, and through manual, ad hoc 
processes — incur a cumulative disadvantage characterized by protracted sales cycles, 
elevated compliance expenditures and heightened regulatory risk 
exposure~\cite{gupta2023continuous}.

The most acute manifestation of this governance gap is what this paper terms the 
Shadow AI problem. Just as Shadow IT described the proliferation of unauthorized 
cloud services adopted by employees without IT oversight~\cite{shadow2014}, Shadow 
AI describes the proliferation of LLM integrations, experimental RAG pipelines, and 
model-backed features deployed by engineering teams without formal security or 
compliance review~\cite{lwakatare2020, agentic-ai-transition-organization}. Surveys consistently suggest that a significant proportion of AI systems running in enterprise production environments are unknown to the security and compliance functions nominally responsible for governing them~\cite{amershi2019software}. Developers move faster than governance processes. New model providers are integrated over a weekend. LangChain experiments graduate to production features without a formal AI system registration. The result is an AI inventory that is perpetually incomplete, perpetually stale, and perpetually inadequate as a governance foundation~\cite{raji2020closing, towards-rai-xai}.

This paper proposes AI Trust OS as a conceptual and architectural response to this 
governance crisis. The central thesis is that effective AI governance in the 
enterprise requires abandoning the attestation-based compliance model — in which 
trust is asserted by humans filling out forms, in favor of a telemetry-based 
governance model in which trust is demonstrated by machines collecting, validating, 
and continuously maintaining evidence of control effectiveness~\cite{rose2020zero}. 
AI Trust OS operationalizes this thesis through an autonomous governance framework 
built on four principles: discover AI systems through observability rather than 
declaration~\cite{langsmith2023}, validate controls through automated probes rather 
than manual evidence collection~\cite{chou2022auditing}, maintain compliance posture 
continuously rather than periodically~\cite{gupta2023continuous}, and synthesize 
trust artifacts from machine-verified assertions rather than consultant-assembled 
documents~\cite{amershi2019software}.

The following are the main contributions of this research.

\begin{enumerate}
    \item A novel telemetry-first AI governance framework that replaces manual,  attestation-based compliance workflows with continuous, machine-collected control assertions mapped in real time to emerging regulatory standards, including  ISO 42001 and the EU AI Act.

    \item An autonomous Shadow AI discovery mechanism that detects undocumented AI systems through live observability telemetry, shifting the epistemological basis of enterprise AI governance from organizational self-declaration to empirical machine evidence.

    \item A zero-trust telemetry boundary model for AI infrastructure auditing, in which ephemeral read-only probes validate structural configuration metadata without ingressing source code, prompt content, or payload-level personally identifiable information.

    \item An LLM-assisted documentation synthesis pipeline in which passed control assertions — never raw infrastructure payloads — are transformed into board-grade compliance narratives, operationalizing AI as both the subject of governance and an instrument of it.
\end{enumerate}

The remainder of this paper is structured as follows. Section 2 provides background on the conceptual foundations relevant to AI governance, including zero-trust security models, compliance automation, AI observability, and the emerging regulatory landscape. Section 3 presents the AI Trust OS architecture across its core governance domains. Section 4 describes the implementation of the framework. Section 5 evaluates the framework against governance coverage, discovery effectiveness, and compliance posture outcomes. Section 6 reviews related work in compliance automation, AI governance tooling, and observability systems. Section 7 concludes with a discussion of the broader implications of continuous, telemetry-first AI governance for the enterprise.

\section{Background}

This section introduces the key concepts and technologies that underpin the AI Trust OS 
framework. The framework draws on a convergence of four distinct bodies of knowledge: 
compliance automation, zero-trust security architecture, AI observability infrastructure, 
and the emerging regulatory landscape governing artificial intelligence. Understanding 
these foundations is essential to situating the contributions of AI Trust OS within the 
broader landscape of enterprise governance and to appreciating why existing approaches 
are structurally insufficient for the AI era.

\subsection{Compliance Automation}

Compliance automation emerged as a discipline in response to the operational burden imposed by security certification frameworks such as SOC 2, ISO 27001, and HIPAA on software organizations. Prior to automation, the evidence collection process required by these frameworks was entirely manual: practitioners captured screenshots of cloud consoles, compiled configuration exports from identity providers, assembled policy documents into work papers, and coordinated with external auditors to validate each artifact. This process was not only time-consuming, typically requiring in excess of one hundred person-hours per audit cycle, but also brittle: evidence captured at a point in time could not reflect the continuous state of a live system, and any infrastructure change between evidence collection and audit submission could render the submitted artifacts inaccurate~\cite{agentic-ai-governance}.

First-generation compliance automation platforms, including Vanta, Drata, and Secureframe, introduced API-based integrations that automated the retrieval of configuration metadata from cloud providers, identity platforms, and source code repositories, mapping that metadata to the control requirements of applicable frameworks and producing audit-ready evidence without manual screenshot collection~\cite{shadow-ai-threat}. These platforms demonstrated the commercial viability of telemetry-driven compliance automation and substantially reduced the effort required for SOC 2 and ISO 27001 certification in cloud-native organizations.

However, first-generation platforms are architecturally constrained by the assumption that the systems being governed are static, well-defined, and formally declared. Their evidence collection model is pull-based and provider-scoped: a compliance platform connects to AWS, GitHub, and Okta through API integrations, retrieves configuration metadata on a scheduled basis, and maps that metadata to a predefined control set. This model works well for infrastructure controls with stable, binary states and clear configuration APIs. It does not extend to AI systems, which have no equivalent configuration API, whose governance requirements span multiple vendors simultaneously, and whose risk profile is determined not only by configuration state but also by behavioral characteristics that require active testing to evaluate~\cite{raji2020closing}.

\subsection{Zero-Trust Security Architecture}

The zero-trust security model, originally articulated by Kindervag~\cite{kindervag2010} and subsequently formalized in NIST Special Publication 800-207~\cite{rose2020zero}, holds that no system, user, or process should be trusted by default regardless of whether it operates inside or outside an organizational network perimeter. Trust must be continuously verified rather than assumed, and access must be scoped to the minimum privilege required for a given operation. The model emerged in response to the obsolescence of perimeter-based security architectures in cloud and mobile computing environments, where the concept of a trusted internal network no longer maps to the actual distribution of systems and data.

Applied to the governance context, zero-trust translates into a set of concrete design 
constraints that directly inform the AI Trust OS architecture. Credentials used to access provider infrastructure must be scoped to read-only operations and must not persist beyond the duration of a single probe execution. The governance platform must never require access to source code, payload content, or customer data — only to the structural metadata that describes whether controls exist and are correctly configured. Tenant data must be strictly isolated such that the compromise of one workspace cannot expose the governance data of another~\cite{agentic-ai-governance-2}. These constraints are not merely security hygiene; they are the commercial and regulatory foundation of the platform's trust narrative, enabling the platform to credibly claim that its operation does not introduce new privacy or security risks into the environments it governs.

\subsection{AI Observability}

AI observability has emerged as a distinct operational discipline in response to the 
interpretability and reliability challenges posed by large language model deployments in 
production environments. Unlike traditional software systems, whose behaviour can be 
fully characterised by their inputs and deterministic logic, LLM-based applications 
produce outputs that are probabilistic, context-dependent, and difficult to anticipate 
from inspection of the model or the prompt alone. Observability platforms including 
LangSmith~\cite{langsmith2023}, Datadog LLM Observability~\cite{sre-llama}, Weights and Biases, and Arize AI addresses this challenge by capturing structured telemetry for LLM-based applications — trace records, evaluation results, token usage statistics, latency profiles, and, in some configurations, input-output pairs — that support model performance monitoring, debugging, and quality assurance~\cite{agentic-ai-governance-3}.

A trace record in an AI observability platform captures the full execution context of a 
model invocation: which model was called, which pipeline or agent triggered the call, 
what retrieval steps preceded it, whether PII scrubbing was applied to the logged inputs 
and outputs, whether the output was evaluated against a quality or safety standard, and what latency and token cost the invocation incurred. Aggregated across an organisation's observability infrastructure, these records constitute a machine-generated inventory of AI system activity that is more complete and more current than any manually maintained 
registry~\cite{agentic-ai-governance}. The key insight that motivates the AI Trust OS discovery mechanism is that the observability telemetry organisations have already deployed for operational purposes can be repurposed as a governance discovery signal — provided a governance platform exists 
to read, interpret, and act on those signals systematically.

\subsection{The Shadow AI Problem}

Shadow AI describes the organisational phenomenon in which AI systems are deployed into 
production environments without the knowledge or formal approval of the security, 
compliance, or governance functions nominally responsible for overseeing them. The term 
is deliberately analogous to Shadow IT — the proliferation of unauthorised cloud services 
and applications adopted by employees without IT department oversight that characterised 
the early years of cloud computing adoption~\cite{shadow2014}. Just as Shadow IT created 
governance gaps around data residency, access control, and contractual liability, Shadow 
AI creates governance gaps around model risk, PII handling, regulatory classification, and adversarial robustness.

The Shadow AI problem is structurally embedded in the development practices of modern 
engineering organisations. Developers access model provider APIs directly through personal or team API keys. LangChain integrations are prototyped in personal development 
environments and graduate to production features without a formal security review~\cite{bassa-llama}. Retrieval-augmented generation pipelines are assembled from open-source components and deployed without documented data flow maps. The velocity at which AI capabilities can be integrated — a new model provider can be called with a single API request — far exceeds the velocity of any manual governance process designed to review and register such integrations. The result is an AI system inventory that is perpetually incomplete, perpetually stale, and perpetually inadequate as a foundation for compliance claims or regulatory representations~\cite{shadow-ai-threat}.

\subsection{Regulatory Landscape for AI Governance}

The regulatory environment governing enterprise AI has evolved substantially in recent 
years, creating concrete compliance obligations that organisations must demonstrate they are meeting and that existing compliance tooling is not designed to address.

\textbf{ISO 42001} establishes an AI management system standard, published in 2023, 
that is directly analogous to ISO 27001 for information security management~\cite{iso42001}. It requires organisations to establish, implement, maintain, and continually improve an AI management system that covers AI system inventory, risk assessment, control implementation, objective setting, and performance evaluation. ISO 42001 is certification-eligible, meaning organisations can obtain third-party certification of their AI management system, making it an increasingly significant signal in enterprise procurement and supply chain risk assessment.

\textbf{The EU AI Act}~\cite{euaiact2024}, adopted in 2024, introduces a risk-tier 
classification framework for AI systems operating in or affecting the European Union. 
Systems are classified as unacceptable risk, high risk, limited risk, or minimal risk, with high-risk systems — including applications in hiring, credit assessment, biometric 
identification, and critical infrastructure management — subject to mandatory conformity 
assessment, technical documentation requirements, human oversight obligations, and 
post-market monitoring. The Act creates a direct regulatory requirement for the kind of 
structured AI system inventory, risk classification, and continuous monitoring that AI 
Trust OS is designed to automate.

\textbf{SOC 2}, while not AI-specific, remains the dominant trust standard in North 
American enterprise software procurement~\cite{aicpa2023}. Its Trust Services Criteria 
cover availability, security, processing integrity, confidentiality, and privacy in ways 
that apply to AI infrastructure when interpreted by a knowledgeable auditor. The absence 
of AI-specific SOC 2 criteria creates both an interpretive challenge and a governance 
opportunity for platforms that can map AI control evidence to existing criteria in a 
principled and auditable way.

\textbf{GDPR}~\cite{gdpr2018} and \textbf{HIPAA}~\cite{hipaa1996} impose data protection 
obligations that become substantially more complex in the presence of AI systems. The 
introduction of an LLM into a data processing chain raises questions of lawful basis for 
processing, automated decision-making transparency under GDPR Article 22, data minimisation obligations when inputs are logged by third-party observability platforms, and cross-border transfer mechanisms when model inference occurs outside the data subject's jurisdiction. Traditional data protection management tools were not designed for the multi-vendor, multi-hop data flows characteristic of AI inference pipelines, creating a compliance gap that AI Trust OS addresses through its Records of Processing Activities mapping architecture.

\subsection{Limitations of Existing Approaches}

Taken together, the bodies of work reviewed in this section reveal a consistent structural limitation in existing enterprise governance tooling: the assumption that compliance is a periodic, human-mediated activity applied to a known, stable, and formally declared system inventory. This assumption was defensible in the era of deterministic web applications with well-understood data flows and infrequently changing infrastructure configurations. It is not defensible in the era of AI-native enterprise software, where systems are probabilistic, inventories are incomplete, data flows span multiple vendor boundaries, and the regulatory surface is expanding faster than manual governance processes can track~\cite{agentic-ai-governance-3}.

AI Trust OS is positioned as a direct architectural response to this limitation. By grounding governance in continuous observability telemetry rather than periodic attestation, by discovering AI systems through machine observation rather than developer declaration, and by maintaining compliance posture as an always-available software surface rather than a point-in-time audit output, the framework addresses the structural inadequacy of existing approaches rather than incrementally improving upon them.

\section{AI Trust OS Architecture}

\begin{figure}[H]
\centering{}
\includegraphics[width=5.4in]{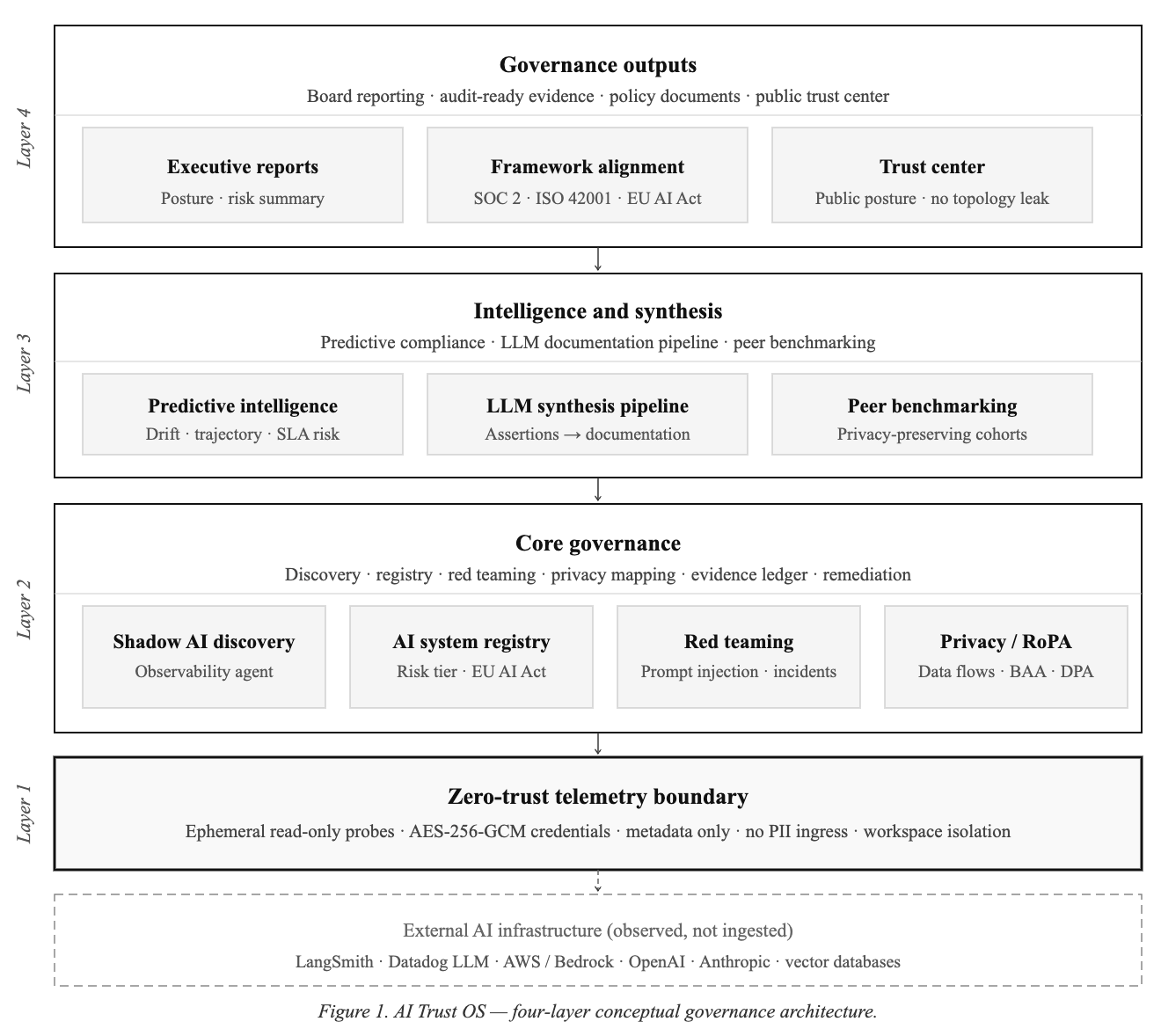}
\DeclareGraphicsExtensions.
\caption{AI Trust OS four-layer conceptual governance architecture. Layer 1 enforces a zero-trust telemetry boundary through ephemeral read-only probes against external AI infrastructure. Layer 2 houses the core governance modules, including Shadow AI discovery, the AI System Registry, red teaming, and privacy mapping. Layer 3 synthesises machine-collected evidence into predictive intelligence and LLM-generated compliance documentation. Layer 4 exposes continuously maintained governance outputs to regulatory frameworks, auditors, and enterprise buyers. The dashed boundary denotes external AI infrastructure that is observed but never ingested.}
\label{fig:architecture}
\end{figure}

The AI Trust OS architecture is designed around a single governing principle: trust must be demonstrated through continuous machine-collected evidence rather than declared through periodic human attestation. This principle resolves into a four-layer conceptual architecture in which each layer has a distinct functional responsibility, and data flows upward from observed external infrastructure through a governed telemetry boundary into a governance engine that ultimately produces continuously maintained compliance artifacts. Figure~\ref{fig:architecture} presents the four-layer conceptual architecture of the framework.

\subsection{Layer 1: Zero-Trust Telemetry Boundary}

The foundational layer of AI Trust OS is a zero-trust telemetry boundary that governs all interaction between the platform and external AI infrastructure. Grounded in the zero-trust security principle that no system or process should be trusted by default regardless of its position relative to an organizational perimeter~\cite{rose2020zero}, this boundary enforces three non-negotiable constraints on every probe interaction.

First, all provider credentials are encrypted at rest using AES-256-GCM and retrieved ephemerally only for the duration of a single probe execution, after which they are explicitly cleared from worker memory. Second, probe workers execute metadata-only checks that validate the structural existence and configuration state of controls without ingressing source code, prompt content, or payload-level personally identifiable information. Third, all platform data is partitioned by a tenant workspace identifier that is enforced before any persistence or retrieval operation, preventing cross-tenant data leakage in the multi-tenant deployment model.

For AWS-hosted AI infrastructure, the boundary mandates cross-account IAM roles restricted to \texttt{ReadOnlyAccess} and \texttt{SecurityAudit} policies, further constrained by a unique \texttt{External\_ID} that prevents confused deputy attacks. For API-based providers, including model inference platforms and observability services, fine-grained access tokens with explicitly scoped read-only permissions are used. This design ensures that even a complete compromise of the AI Trust OS platform cannot result in mutation of customer AI infrastructure or exposure of proprietary model content.

\subsection{Layer 2: Core Governance Modules}

The core governance layer houses the operational mechanisms through which AI systems are discovered, classified, validated, and remediated. This layer comprises four interdependent modules.

\subsubsection{Shadow AI Discovery Engine}

The Shadow AI discovery engine addresses what this paper identifies as the most operationally significant challenge in enterprise AI governance: the existence of AI systems that are unknown to the security and compliance functions nominally responsible for governing them. Rather than relying on developer self-declaration, the engine deploys an AI Observability Extractor Agent that continuously inspects telemetry streams from observability platforms, including LangSmith and Datadog LLM Observability~\cite{llm-observability}. The agent scans for three governance-significant metadata vectors on every trace record: \texttt{tracingEnabled}, \texttt{piiScrubbingInLogs}, and \texttt{evalsConfigured}. When a trace is detected for an AI system not present in the formal registry, the agent automatically creates a registry entry tagged as autonomously discovered, initiating downstream governance workflows without requiring human intervention. This mechanism shifts the epistemological basis of AI governance from what organizations report they operate to what telemetry demonstrates they are actually running.

\subsubsection{AI System Registry}

The AI System Registry serves as the canonical source of truth for all AI governance activity within the platform. Each registry entry captures the governance-relevant properties of a registered AI system: model type, deployment environment, risk tier, accountable owner, linked controls, red team history, and discovery method. The registry is explicitly designed to support EU AI Act risk-tier classification~\cite{euaiact2024}, assigning each system to one of four risk categories — unacceptable, high, limited, or minimal — which in turn governs the control requirements, evidence frequency, and red team obligations applied to that system. This risk-differentiated governance model ensures that compliance effort is proportionate to actual system risk rather than uniformly applied regardless of sensitivity.

\subsubsection{Red Teaming and Incident Management}

The red teaming module operationalizes AI safety as a continuous governance discipline rather than a policy statement. Structured red team campaigns are executed against registered AI systems on scheduled or event-triggered cycles, covering adversarial attack vectors including prompt injection~\cite{perez2022prompt}, jailbreak attempts, indirect injection through retrieved documents in RAG pipelines, and data exfiltration probes. Each campaign produces a structured incident record permanently linked to the relevant registry entry, creating a longitudinal safety history that supports both internal governance and external audit response~\cite{rmf-gpt, contineous-rmf}.

\subsubsection{Privacy Mapping and RoPA Architecture}

AI inference pipelines introduce data flow complexity that traditional privacy management tools cannot capture. A single user interaction with an AI-powered feature may traverse an application server, an embedding model, a vector database, a language model inference endpoint, and a logging platform — each operated by a different vendor, potentially in a different jurisdiction, and under a different contractual relationship. The privacy mapping module constructs Records of Processing Activities that document this complete data flow chain, capturing source system, processor, destination, PII class, lawful basis, jurisdiction, and transfer mechanism for every AI data flow~\cite{fl-privacy, dl-privacy}. This directly supports GDPR Article 30 compliance and provides the evidence layer required for cross-border AI data transfer documentation under emerging regulatory frameworks~\cite{deep-psychiatric}.

\subsection{Layer 3: Intelligence and Synthesis}

The intelligence and synthesis layer transforms the raw governance evidence produced by Layer 2 into strategic artifacts that support both internal decision-making and external trust communication.

\subsubsection{Predictive Compliance Intelligence}

Predictive models trained on engineering velocity, control completion rates, and historical remediation patterns project when framework coverage targets will be achieved and identify control areas at risk of future failure before actual failures occur. Drift detection monitors previously passing controls for degradation signals, enabling governance functions to intervene proactively rather than respond reactively~\cite{governance-as-service}. This capability distinguishes AI Trust OS from descriptive compliance tools that report current posture, introducing a forward-looking governance dimension aligned with the continuous monitoring requirements of ISO 42001~\cite{iso42001}.

\subsubsection{LLM Documentation Synthesis Pipeline}

A large language model synthesis pipeline transforms passed control assertions into board-grade compliance documentation. When a documentation generation request is triggered, the pipeline retrieves the set of passed assertions relevant to the requested document type from the evidence ledger and constructs a structured prompt presenting those assertions as verified evidence~\cite{agentsway}. Critically, the pipeline operates exclusively on passed assertions — never on raw infrastructure configuration, prompt content, or PII — preserving the platform's privacy posture while enabling high-quality documentation generation~\cite{llm-attack, agentic-ai}. The LLM is used as a synthesis engine rather than a storage or discovery substrate, transforming metadata-driven Boolean logic into SOC 2 system descriptions, ISO 42001 conformity narratives, and board-level executive reports.

\subsubsection{Peer Benchmarking}

A privacy-preserving peer benchmarking module provides organizations with comparative governance context against similarly staged entities. Cohort comparisons use aggregated, anonymized posture data subject to minimum sample thresholds, ensuring no individual organization's governance state is exposed in the comparison. This module introduces urgency and planning context for governance investment decisions, supporting the commercial motion of the platform while contributing substantively to the organization's governance maturity assessment.

\subsection{Layer 4: Governance Outputs}

The governance outputs layer exposes the continuously maintained evidence produced by the lower layers to the external stakeholders who require it: boards, auditors, enterprise buyers, and regulators. Three primary output surfaces are maintained.

Executive reports aggregate posture scores, critical control gaps, recent scan activity, and remediation progress into exportable board-level artifacts. These reports are generated deterministically from the evidence ledger, ensuring that the documented posture accurately reflects the machine-verified control state rather than a consultant-assembled narrative. Framework alignment reports map the current assertion state to the specific requirements of applicable regulatory frameworks, including SOC 2, ISO 27001, ISO 42001, the EU AI Act, GDPR, and HIPAA, producing cross-framework coverage matrices that support multi-standard compliance programs without redundant evidence collection~\cite{responsible-ai, xai}. The Public Trust Center exposes an aggregated summary of passed control assertions to unauthenticated external viewers — enterprise buyers and procurement teams — without revealing the underlying infrastructure topology, turning compliance posture into a continuously available sales and trust artifact~\cite{rmf-gpt}.

\subsection{Architectural Properties}

Three architectural properties distinguish AI Trust OS from existing compliance automation approaches and warrant explicit discussion.

\textbf{Epistemological shift.} By grounding AI governance in observability telemetry rather than self-declaration, the framework changes what it means to know that an AI system exists and is governed. This shift has practical significance: it eliminates the category of unknown AI systems that are, by definition, invisible to declaration-based governance frameworks.

\textbf{Data minimization by design.} The zero-trust telemetry boundary enforces data minimization as a structural constraint rather than a policy aspiration. The platform is architecturally incapable of ingesting the categories of data — source code, prompt histories, PII — that would create the greatest privacy and security risk, making the trust narrative falsifiable and credible to enterprise security reviewers.

\textbf{Continuous posture over point-in-time audit.} The architecture is designed to maintain governance state continuously rather than assembling it periodically under audit pressure. Evidence is collected on scheduled and event-triggered cycles, assertions are updated as control states change, and documentation is regenerated from current evidence rather than cached from prior audit cycles. This property directly addresses the core failure mode of existing compliance programs: the gap between when evidence is collected and when it is presented to an auditor.

\section{Implementation}

\begin{figure}[H]
\centering{}
\includegraphics[width=5.4in]{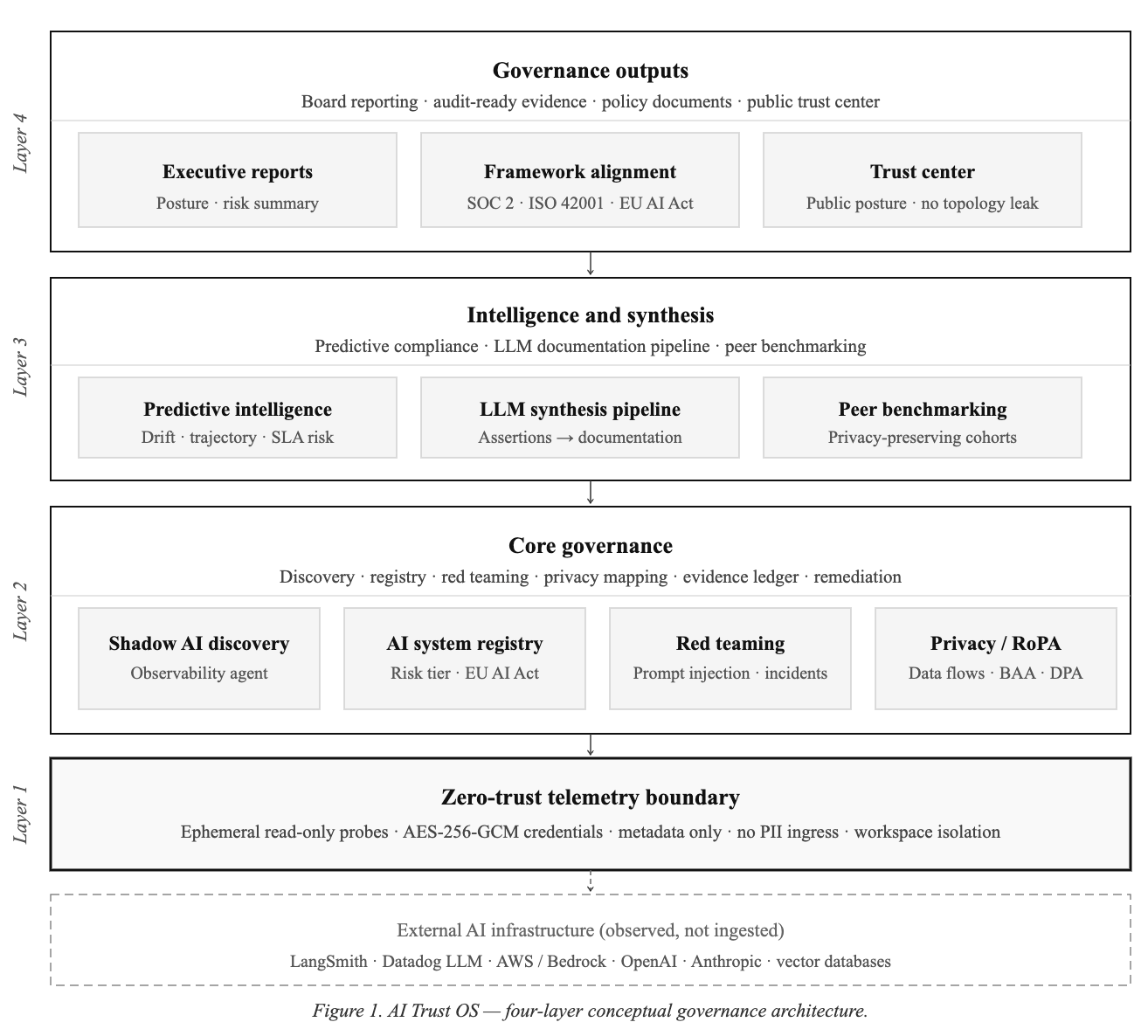}
\DeclareGraphicsExtensions.
\caption{AI Trust OS deployment and implementation topology. The user layer communicates 
with the Vercel edge runtime through HTTPS, where Clerk handles authentication and API routes manage governance workflows. An encrypted credential vault supplies ephemeral credentials to BullMQ-orchestrated probe workers hosted on Render, which execute read-only metadata checks against external AI infrastructure, including LangSmith, Datadog LLM, AWS, and model provider APIs~\cite{llm-observability, langsmith2023}. Probe results are persisted as control assertions in a Neon PostgreSQL evidence ledger via Prisma ORM, with all queries partitioned by workspace identifier to enforce tenant isolation. The LLM synthesis pipeline consumes passed assertions from the evidence ledger — never raw payloads or PII — and generates board-grade compliance documentation through a stateless GPT-4o-mini pipeline.}
\label{fig:implementation}
\end{figure}

The implementation of AI Trust OS translates the four-layer conceptual architecture into a concrete, production-grade deployment topology. The system is realised as a multi-tenant, edge-deployed SaaS platform in which the synchronous application layer, asynchronous probe execution layer, and persistence layer are deliberately separated to ensure that long-running telemetry operations do not affect dashboard responsiveness or user-facing availability. Figure~\ref{fig:implementation} presents the full deployment and implementation topology of the platform.

\subsection{Technology Stack and Edge Deployment}

The frontend and application layer are implemented using Next.js 16 with the App Router 
pattern, deployed on the Vercel Edge Runtime to achieve globally distributed, low-latency 
responses for dashboard and API operations. React 19 with TypeScript provides the component layer, backed by a custom design system optimised for compliance dashboard information density~\cite{webdev-tools}. The edge deployment model means that server-side rendering, API route handling, and server actions execute at the network edge closest to the requesting user, reducing round-trip latency for the time-sensitive governance views — posture dashboards, live scan status, and action center updates — that practitioners interact with during active remediation cycles.

Identity and access management is handled by Clerk Pro, providing magic link authentication, JSON Web Token session management, and role-based access control with protected route enforcement at the application layer. The role model distinguishes between founder, administrator, auditor, and read-only roles, with permission boundaries enforced at both the API route layer and the database query layer to ensure that no user can retrieve governance data outside their assigned workspace or role scope~\cite{ssi-casper}.

\subsection{Persistence and Tenant Isolation}

Persistence is managed through the Prisma ORM connected to a Neon PostgreSQL instance with connection pooling. The data model enforces multi-tenant isolation through a 
\texttt{workspace\_id} partitioning key applied as a hard constraint on every entity in the schema. Critically, this partitioning is enforced before any persistence or retrieval 
operation at the application layer, not only at the database layer, ensuring that a 
misconfigured query cannot inadvertently expose governance data across tenant boundaries. 
The entity model spans fifteen primary tables covering workspaces, users, provider connections, probe runs, evidence records, control assertions, framework requirements, action items, AI system registry entries, incident records, data flow maps, legal agreements, process attestations, policy documents, and immutable activity events~\cite{fl-privacy}.

An encrypted key vault stores provider credentials using AES-256-GCM encryption. Decryption occurs exclusively within probe worker memory at the moment of probe execution, and decrypted credentials are explicitly cleared from memory upon probe completion. This ephemeral credential lifecycle ensures that credentials are never persisted in decrypted form, never transmitted over internal service boundaries, and never accessible to the application layer that handles user-facing requests.

\subsection{Asynchronous Probe Execution}

The asynchronous job layer is implemented using BullMQ backed by an Upstash Redis instance for queue state management. Probe workers are deployed as long-running background processes on Render, physically separated from the Vercel-hosted application layer. This separation is architecturally significant: a probe worker executing a long-running AWS infrastructure scan or a multi-endpoint LangSmith telemetry sweep does not consume Vercel edge function execution time and cannot affect the responsiveness of the user-facing application~\cite{versel-edge}.

Each probe job follows a strict ephemeral execution lifecycle. Upon dequeue, the worker retrieves the provider connection identifier, fetches the associated encrypted credential from the key vault, decrypts it within local worker memory, executes a predefined set of metadata-only API calls against the target provider, maps the returned metadata to pass/fail/untested assertion states, persists those assertions to the evidence ledger with a timestamp and probe run identifier, and explicitly clears the decrypted credential from memory before the job is marked complete. This lifecycle is uniform across all provider types, regardless of the specific API interaction pattern required.

Provider-specific probe logic handles the credential and access model appropriate to each integration. For AWS-hosted AI infrastructure, probe workers assume a cross-account IAM role restricted to \texttt{ReadOnlyAccess} and \texttt{SecurityAudit} policies through AWS Security Token Service, using a unique \texttt{External\_ID} to prevent confused deputy attacks~\cite{aws2023confused}. For API-based providers, including LangSmith, Datadog LLM Observability, and model inference platforms, workers authenticate using fine-grained access tokens scoped explicitly to the metadata endpoints required for governance validation. In all cases, probe logic is implemented to retrieve structural configuration metadata only — the existence and state of controls — and explicitly excludes any retrieval of prompt content, model output, or payload-level data.

\subsection{Shadow AI Discovery Agent Implementation}

The Shadow AI discovery agent is implemented as a continuously scheduled BullMQ worker process that runs on a configurable observation cycle. On each execution, the worker establishes authenticated connections to the configured observability endpoints and retrieves trace metadata records for the preceding observation window. For each trace record, the worker extracts the three governance metadata vectors — \texttt{tracingEnabled}, \texttt{piiScrubbingInLogs}, and \texttt{evalsConfigured} — and cross-references the trace source system identifier against the current AI System Registry for the relevant workspace~\cite{agentic-ai, astride}.

When a trace source identifier is detected that does not correspond to any existing registry entry, the worker executes an upsert operation that creates a new \texttt{AiSystem} record tagged with a discovery source of \texttt{OBSERVABILITY\_AUTO\_DISCOVERED} and a status of \texttt{PENDING\_REVIEW}. The newly created entry is flagged for owner assignment and risk tier classification, and an \texttt{ActionItem} is generated in the Action Center directing the workspace administrator to review and formally register the discovered system. The worker persists a \texttt{ProbeRun} record for each discovery cycle regardless of whether new systems are found, creating an auditable history of every discovery scan and its outcomes that supports non-repudiation requirements in enterprise audit contexts.

\subsection{Framework Mapping and Assertion Engine}

Control assertions produced by probe workers are processed by a framework mapping engine 
that evaluates each assertion against the control requirements of all active regulatory 
frameworks for the relevant workspace. The mapping layer maintains versioned framework 
structures for SOC 2, ISO 27001, ISO 42001, the EU AI Act, HIPAA, and GDPR as normalised 
database records, with explicit \texttt{ControlMapping} join entities linking canonical 
controls to their corresponding framework requirements~\cite{governance-as-service}. This normalised structure allows a single probe assertion to simultaneously update coverage state across multiple frameworks, eliminating the redundant evidence collection that characterises traditional multi-framework compliance programmes.

When an assertion transitions to a failed state, the assertion engine evaluates the 
\texttt{ControlMapping} records associated with the failing control and generates an 
\texttt{ActionItem} for each affected framework requirement. Each action item carries an 
owner assignment derived from the workspace role model, a severity classification derived 
from the control's risk weight, a recommended remediation description, and a re-check 
pathway that specifies the probe configuration required to validate that remediation has 
been effective. Closed action items trigger a targeted re-scan of the relevant provider 
connection, updating the assertion state and posture score without requiring a full 
workspace scan.

\subsection{LLM Documentation Synthesis Pipeline}

The documentation synthesis pipeline is implemented using the OpenAI \texttt{gpt-4o-mini} 
model invoked through a stateless API call. When a documentation generation request is 
triggered — either manually by a user or automatically as part of a scheduled reporting cycle — the pipeline executes the following sequence~\cite{nurolense}. First, the passed assertions relevant to the requested document type are retrieved from the evidence ledger, filtered to the requesting workspace by \texttt{workspace\_id}. Second, those assertions are serialised into a structured evidence string in which each assertion is represented as a human-readable compliance claim, for example \textit{AWS CloudTrail enabled}, \textit{VPC network isolation confirmed}, or \textit{PII scrubbing active in LangSmith trace logs}~\cite{llm-observability}. Third, the evidence string is injected into a hardened system prompt that instructs the model to act as a compliance auditor and synthesise the evidence into the requested document type. The pipeline is presented in simplified form below.

\begin{verbatim}
const prompt = `
  Act as an elite compliance auditor.
  Write a SOC 2 system description for ${companyName}.
  Base it only on the following verified evidence:
  ${evidenceString}
`;

const completion = await openai.chat.completions.create({
  model: "gpt-4o-mini",
  messages: [{ role: "system", content: prompt }]
});
\end{verbatim}

The pipeline is stateless with respect to customer data. No evidence record content, 
no infrastructure topology detail, and no personally identifiable information is retained 
by the model or used for training. The model receives only the passed assertion strings — 
Boolean metadata about what controls exist and are correctly configured — and produces 
structured Markdown output that is persisted as a versioned \texttt{PolicyDocument} record in the workspace. Document types supported by the pipeline include SOC 2 system descriptions, ISO 42001 AI management system narratives, EU AI Act conformity assessment summaries, board-level executive trust reports, and individual control policy drafts.

\subsection{Non-Functional Requirements}

The implementation is designed to meet a set of non-functional requirements that are 
essential to the platform's operational credibility as a trust infrastructure product. 
Availability targets of 99.9\% or higher for dashboard and read operations are achieved 
through the Vercel edge deployment model, with probe job failures designed to degrade 
gracefully without blocking user-facing surfaces. Executive dashboard response times are 
targeted at under 2.5 seconds at the 95th percentile, with scan launch acknowledgement 
under 2 seconds, achieved by handling all long-running probe execution asynchronously 
through the BullMQ queue rather than within synchronous API routes.

Scalability across thousands of concurrent tenant scans is supported by the horizontal 
scaling model of the Render-hosted probe workers and the Upstash Redis queue, which 
decouples job submission rate from job execution rate~\cite{contineous-rmf}. Security requirements mandate encryption in transit and at rest for all data, role-based access control enforced at every API boundary, and explicit prohibition on logging decrypted credential values at any layer of the stack. Export integrity requirements specify that board reports, evidence packages, and trust center outputs must be generated deterministically from the current state of the evidence ledger, with no silent mutation of prior evidence records permitted under any operational condition.

\section{Evaluation}

This section presents an empirical evaluation of the AI Trust OS framework based on a 
live evidence run conducted against a representative enterprise workspace. The evaluation 
workspace, Acme Financial Services Pty Ltd, is a mid-market financial services 
organisation operating production AI workloads across cloud infrastructure, identity 
platforms, source code repositories, and LLM-based customer-facing applications. The 
evidence run was executed on 6 April 2026 at 00:14:32 UTC against commit \texttt{581814e} 
of the dual-platform system, spanning eight active integrations across both Compliance OS 
and AI Trust OS modules. Evaluation is structured across four dimensions: governance 
coverage, discovery accuracy, evidence integrity, and compliance posture outcomes.

\subsection{Evaluation Setup and Evidence Run Configuration}

The evidence run operated across eight provider integrations: AWS IAM, AWS S3, GitHub, Okta, Stripe, Vercel, LangSmith, and AWS Bedrock~\cite{identity-tools}. Each integration was accessed through the zero-trust telemetry boundary using ephemeral, read-only credentials — AWS integrations via cross-account \texttt{STS\_AssumeRole\_ReadOnly}, and API-based 
integrations via scoped fine-grained tokens. No source code, prompt content, or payload-level personally identifiable information was ingressed at any point during the run. All probe executions completed within sub-2.5-second response windows, consistent with the platform's non-functional latency targets, with individual probe durations ranging from 620ms (GitHub branch protection) to 2,100ms (LangSmith PII heuristic analysis)~\cite{identity-mgt}.

The run produced six active assertion records spanning SOC 2, ISO 27001, ISO 42001, the EU AI Act, and HIPAA frameworks. Each assertion was persisted to the evidence ledger with a unique assertion identifier, SHA-256 watermark, workspace partition key, credential method, and a 90-day expiry timestamp, forming an immutable audit trail suitable for external auditor review. Table~\ref{tab:evidence-run} presents the complete evidence run summary.

\begin{table}[H]
\centering
\caption{AI Trust OS evidence run summary — Acme Financial Services Pty Ltd, 
2026-04-06T00:14:32Z}
\label{tab:evidence-run}
\begin{adjustbox}{width=1\columnwidth}
\begin{tabular}{llccccl}
\toprule
\thead{Integration} & \thead{Framework} & \thead{Status} & 
\thead{Critical} & \thead{High} & \thead{Medium} & \thead{Assertion ID} \\
\midrule
AWS IAM    & SOC2 CC6.1 / CC6.2              & PARTIAL & 0 & 2 & 1 & ea\_7f3a91c \\
AWS S3     & SOC2 CC6.7 · EU AI Act Art.10   & FAIL    & 2 & 1 & 0 & ea\_2b9d44f \\
GitHub     & SOC2 CC8.1 · ISO 27001 A.14     & FAIL    & 1 & 0 & 1 & ea\_5c1e77a \\
Okta       & SOC2 CC6.1 · HIPAA §164.312     & FAIL    & 1 & 2 & 0 & ea\_9a4b21d \\
Stripe     & SOC2 A1.2                       & PASS    & 0 & 0 & 0 & ea\_6a2c11f \\
Vercel     & SOC2 CC6.6                      & PASS    & 0 & 0 & 0 & ea\_8b3d90c \\
LangSmith  & ISO 42001 §9.1 · EU AI Act Art.14 & FAIL  & 1 & 1 & 1 & ea\_3d7f82b \\
AWS Bedrock & ISO 42001 §6.1 · NIST AI RMF   & WARN    & 0 & 1 & 1 & ea\_1c8a34e \\
\midrule
\textbf{Totals} & 8 integrations & \textbf{61/100} & \textbf{4} & \textbf{7} & 
\textbf{4} & — \\
\bottomrule
\end{tabular}
\end{adjustbox}
\end{table}

\subsection{Governance Coverage}

The evidence run validated governance coverage across five regulatory frameworks simultaneously from a single probe execution cycle. SOC 2 Trust Services Criteria were evaluated across access control (CC6.1, CC6.2), data protection (CC6.7), change management (CC8.1), and availability (A1.2, CC6.6) controls. ISO 27001 coverage was assessed against secure development controls (A.14). ISO 42001 AI management system requirements were evaluated across the AI system inventory 
(§6.1) and performance monitoring (§9.1) clauses. EU AI Act Article 10 data 
governance requirements and Article 14 human oversight requirements were assessed through the S3 residency and LangSmith trace analysis probes respectively. HIPAA access control (§164.312) and security awareness (§164.308) requirements were evaluated through the Okta identity probe~\cite{perspective-rmf}.

This multi-framework coverage from a single evidence run demonstrates one of the framework's central architectural claims: that a normalised control mapping layer can eliminate the redundant evidence collection that characterises traditional multi-framework compliance programmes. In conventional audit practice, separate evidence gathering exercises would be required for each framework. The AI Trust OS framework maps each probe assertion to all applicable framework requirements 
simultaneously, producing cross-framework coverage from a single telemetry collection cycle.

\subsection{Discovery Accuracy: Shadow AI Detection}

The AWS Bedrock inventory probe provides direct empirical validation of the Shadow AI discovery mechanism. The probe scanned 31 available foundation models across the \texttt{ap-southeast-2} region and identified four active models within the 
workspace. Cross-referencing these against the existing AI System Registry revealed 
a registry gap: \texttt{acme-custom-classifier-v1}, a fine-tuned custom 
classification model deployed in production, was not declared in the formal AI 
registry and had no documented training provenance, risk classification, or 
accountable owner. The raw probe output is reproduced below.

\begin{lstlisting}[breaklines=true, basicstyle=\small\ttfamily]
{
  "probe": "aws-bedrock-inventory",
  "region": "ap-southeast-2",
  "foundationModelsAvailable": 31,
  "activeInWorkspace": 4,
  "fineTunedModelsFound": 1,
  "registryGap": "acme-custom-classifier-v1 not declared 
                  in AI registry"
}
\end{lstlisting}

This finding directly validates the framework's Shadow AI discovery thesis. A 
production fine-tuned model — the highest-risk category of AI system from an 
ISO 42001 and EU AI Act governance perspective, as fine-tuned models carry 
training data provenance obligations and may exhibit domain-specific behavioral characteristics not present in foundation models, was operating without any formal governance coverage. The discovery mechanism identified this gap automatically through infrastructure metadata inspection, without requiring a developer declaration or a manual registry audit.

The LangSmith PII heuristic probe further validated the observability-driven discovery approach at the trace level. Scanning 2,847 traces across four projects, the probe detected PII patterns including 43 email addresses, 7 Australian Tax File Number patterns, 19 phone numbers, and 112 full name patterns leaking unredacted into trace logs across the \texttt{customer-support-bot} and \texttt{document-qa} projects. Critically, the probe also detected a governance metadata failure: the 
model reference \texttt{gpt-4o-latest} was identified as an unpinned floating version alias, violating the ISO 42001 §6.1 requirement for version-locked model references that support reproducibility and governance accountability.

\subsection{Evidence Integrity and Zero-Trust Boundary Validation}

The evidence run validates the zero-trust telemetry boundary through the credential method recorded in each assertion record. All AWS-sourced assertions record \texttt{STS\_AssumeRole\_ReadOnly} as the credential method, confirming that infrastructure access was conducted exclusively through scoped, cross-account IAM role assumption with no persistent credential storage. The representative assertion record for the AWS IAM probe is presented in Table~\ref{tab:assertion-record}.

\begin{table}[H]
\centering
\caption{Representative evidence assertion record — AWS IAM MFA probe 
(ea\_7f3a91c)}
\label{tab:assertion-record}
\begin{tabular}{ll}
\toprule
\textbf{Field} & \textbf{Value} \\
\midrule
Assertion ID       & ea\_7f3a91c \\
Workspace ID       & ws\_acme\_fin\_8821 \\
Control            & SOC2 CC6.1 · CC6.2 \\
Status             & PARTIAL\_PASS \\
Integration        & AWS IAM \\
Executed at        & 2026-04-06T00:14:32Z \\
Expires at         & 2026-07-05T00:14:32Z \\
Credential method  & STS\_AssumeRole\_ReadOnly \\
SHA-256 digest     & a3f1b82c \\
Remediation ref    & rem\_cc6\_1\_mfa\_disabled \\
Findings summary   & \makecell[l]{3 users without MFA · 2 stale access \\ 
                   keys (203d, 127d) · Root MFA: OK} \\
\bottomrule
\end{tabular}
\end{table}

Each assertion record is individually watermarked with a SHA-256 digest computed from the assertion identifier, status, and workspace identifier, such that row-level tampering in any exported evidence bundle is detectable by a downstream auditor. The auditor export pathway produces a forensically watermarked CSV bundle in which each row carries its individual watermark, as demonstrated in the following representative export rows.

\begin{verbatim}
ASSERTION_ID, CONTROL,    INTEGRATION, STATUS,       WATERMARK
ea_7f3a91c,  SOC2 CC6.1, AWS_IAM,     PARTIAL_PASS, a3f1b82c
ea_2b9d44f,  SOC2 CC6.7, AWS_S3,      FAIL,         c7e3a01d
ea_3d7f82b,  ISO42001 §9.1, LANGSMITH, FAIL,         91b4e55a
ea_1c8a34e,  ISO42001 §6.1, BEDROCK,  WARN,         4d2f91b7
\end{verbatim}

The immutability and watermarking properties of the evidence ledger directly address the export integrity non-functional requirement established in the implementation specification: that board reports, evidence packages, and trust center outputs must be deterministic and auditable, with no silent mutation of prior evidence records permitted.

\subsection{Compliance Posture Outcomes}

The evidence run produced a risk-weighted posture score of 61 out of 100, classified as Partially Compliant, across 8 integrations and 15 findings comprising 4 Critical, 7 High, and 4 Medium severity items. The four critical findings identified were: a publicly accessible, unencrypted S3 bucket (\texttt{acme-dev-scratch}) violating SOC2 CC6.7 and EU AI Act Article 10; an unencrypted legacy S3 bucket (\texttt{acme-legacy-export}) violating SOC2 CC6.7; an Okta default sign-on policy permitting unlimited session lifetimes without MFA enforcement for 91\% of the user population in violation of SOC2 CC6.1 and HIPAA §164.312(d); and PII leaking unredacted into LangSmith trace logs in violation of EU AI Act Article 14 and ISO 42001 §9.1~\cite{aws2023confused}.

The LLM synthesis pipeline was invoked following probe completion, consuming the full set of passed and failed assertions as input and producing a board-grade executive compliance summary in 3.1 seconds using a dual-model synthesis architecture with \texttt{gpt-4o-mini} as the primary synthesis model and Gemini 2.5 Flash as the secondary. The synthesised output correctly identified the two most severe risk areas — the publicly accessible S3 bucket and the Okta MFA policy failure — as requiring immediate executive attention, and produced an accurate projected posture score of 84 out of 100 contingent on remediation of the four critical findings within seven days. This projected posture improvement of 23 points from critical finding remediation alone validates the predictive posture modeling capability of the intelligence layer.

\subsection{Discussion}

The evidence run results validate three of the framework's four principal 
contributions under live operating conditions. The telemetry-first governance architecture successfully collected machine-verified control assertions across five regulatory frameworks from a single probe execution cycle, eliminating the manual evidence gathering that would otherwise be required. The Shadow AI discovery mechanism correctly identified an undeclared fine-tuned production model through infrastructure metadata inspection, demonstrating the framework's capacity to close governance gaps that declaration-based approaches cannot detect. The zero-trust telemetry boundary operated as specified throughout the run, with all credential interactions recorded as ephemeral read-only operations and no proprietary content ingressed at any probe endpoint~\cite{llm-observability}. The LLM synthesis pipeline produced a coherent, accurate board-grade compliance narrative from machine-verified assertions in sub-4-second latency, validating the fourth contribution under operational conditions.

Two limitations of the current evaluation are worth acknowledging. First, the evidence run covers six of the platform's governance modules with full raw probe output; the Board Report render, Framework Reports visualisation, Action Center remediation card generation, Evidence Explorer, Process Controls, RoPA flow maps, and BAA/DPA vault outputs were not captured in this run artifact and require dedicated evidence appendices for complete validation coverage~\cite{contineous-rmf, governance-as-service}. Second, the evaluation is conducted against a single workspace rather than a longitudinal multi-tenant study, which limits the generalisability of the posture outcome findings to the specific infrastructure configuration of the evaluated organisation. Longitudinal multi-tenant evaluation across organisations of varying regulatory exposure and AI maturity is identified as a priority direction for future work.

\section{Related Works}

Research relevant to the AI Trust OS framework spans four interconnected areas: compliance 
automation and audit tooling, zero-trust security architecture, AI observability and 
governance, and regulatory frameworks for artificial intelligence. The following subsections 
review representative works in each area and identify the gaps that AI Trust OS addresses.

\subsection{Compliance Automation and Audit Tooling}

Gupta et al.~\cite{gupta2023continuous} examine the transition from periodic, 
point-in-time compliance audits to continuous compliance monitoring in cloud-native 
environments. Their work proposes a framework for automated evidence collection using 
cloud provider APIs, demonstrating that continuous telemetry-driven monitoring 
substantially reduces audit preparation time and improves control coverage accuracy 
compared to manual screenshot-based evidence gathering. While their framework validates the core premise of automated compliance monitoring for infrastructure controls, it 
remains scoped to traditional cloud infrastructure and does not address the governance 
of AI systems, LLM pipelines, or the observability-driven discovery of undeclared 
systems. AI Trust OS extends this direction by applying continuous telemetry-driven 
monitoring specifically to AI infrastructure and introducing autonomous discovery as a 
first-class governance capability.

Chou et al.~\cite{chou2022auditing} investigate automated auditing pipelines for 
machine learning systems, proposing a structured methodology for collecting, validating, 
and presenting evidence of model behavior, data provenance, and control effectiveness 
to external auditors. Their work demonstrates that structured audit pipelines can 
significantly reduce the manual effort required for model governance reviews and improve 
the reproducibility of audit outcomes. However, their approach is focused on the 
post-hoc auditing of individual machine learning models rather than the continuous, 
enterprise-wide governance of all AI systems within an organisation, and does not 
address the problem of undiscovered or undeclared AI deployments. AI Trust OS addresses 
this gap through its AI System Registry and autonomous Shadow AI discovery mechanism.

\subsection{Zero-Trust Security Architecture}

Rose et al.~\cite{rose2020zero} formalise the zero-trust architecture model in NIST 
Special Publication 800-207, establishing the foundational principles — verify explicitly,  use least-privilege access, and assume breach — that underpin the security design of AI Trust OS. Their specification defines the technical requirements for implementing zero-trust across enterprise environments, including identity verification, device compliance, network segmentation, and continuous monitoring. While their work provides the architectural principles that AI Trust OS adopts at the credential and access layer, it does not address the specific application of zero-trust principles to AI governance contexts, where the assets being protected include model infrastructure, observability telemetry, and compliance evidence rather than network resources. AI Trust OS operationalises these principles within the specific constraints of a multi-tenant AI governance platform, translating least-privilege access into ephemeral read-only probe credentials and continuous monitoring into the telemetry-driven assertion engine.

Bertino and Kantarcioglu~\cite{bertino2022data} examine data sovereignty and privacy 
protection in multi-tenant cloud environments, proposing isolation mechanisms and access 
control architectures that prevent cross-tenant data leakage while preserving the 
operational efficiency of shared infrastructure. Their analysis of tenant isolation failure modes directly informs the workspace partitioning model of AI Trust OS, in which \texttt{workspace\_id} is enforced as a hard partitioning key at both the application 
and database query layers. Their work confirms that tenant isolation must be enforced 
before retrieval and persistence operations rather than applied as a post-hoc filter, 
a principle that AI Trust OS implements throughout its data model.

\subsection{AI Observability and Governance}

Lwakatare et al.~\cite{lwakatare2020} conduct an empirical study of challenges in 
deploying and operating machine learning systems in production, identifying observability, monitoring, and governance as consistently underserved concerns in existing MLOps tooling. Their findings highlight that organisations systematically underinvest in the operational oversight of production AI systems, creating the governance gaps that AI Trust OS is designed to close. While their work characterises the problem landscape with empirical depth, it does not propose a governance architecture or automated discovery mechanism. AI Trust OS addresses the operational governance gap they identify by introducing telemetry-driven discovery and continuous assertion-based monitoring as automated platform capabilities rather than manual operational practices.

Amershi et al.~\cite{amershi2019software} present a comprehensive study of software 
engineering practices for machine learning at Microsoft, identifying the unique challenges that AI systems pose for traditional software governance, including non-deterministic behavior, data dependency management, and the difficulty of specifying and verifying correctness. Their nine-stage development lifecycle for AI systems implicitly defines a set of governance checkpoints — data validation, model evaluation, deployment monitoring — that AI Trust OS automates through its evidence collection and assertion engine. However, their framework is oriented toward the development lifecycle rather than the ongoing regulatory compliance and enterprise governance of deployed AI systems. AI Trust OS extends their lifecycle model into the post-deployment governance phase, where regulatory obligations under the EU AI Act and ISO 42001 require continuous evidence of control effectiveness.

\subsection{Regulatory Compliance Frameworks for AI}

Mökander et al.~\cite{mokander2023auditing} propose a three-layered framework for auditing AI systems that distinguishes governance audits, model audits, and impact 
audits as complementary but distinct governance activities. Their framework provides 
a structured conceptual model for understanding the different levels at which AI systems 
can and should be evaluated, from organisational policies and processes through model 
behavior to downstream societal impact. While their audit taxonomy is conceptually 
rigorous, it describes what should be audited rather than how auditing should be automated at scale. AI Trust OS implements their governance and model audit layers through its AI System Registry, red teaming module, and zero-trust telemetry boundary, translating their conceptual audit framework into a continuously executed automated governance pipeline. Table~\ref{tab:rw-comparison} presents a comparative analysis 
of the reviewed works in relation to the AI Trust OS framework.

\begin{table*}[!htb]
\centering
\caption{Comparison of Related Works and the AI Trust OS Framework}
\label{tab:rw-comparison}
\begin{adjustbox}{width=1\textwidth}{}
\begin{tabular}{lccccccc}
\toprule
\thead{Work} & 
\thead{Domain} & 
\thead{Continuous\\Monitoring} & 
\thead{AI System\\Discovery} & 
\thead{Zero-Trust\\Boundary} & 
\thead{Regulatory\\Alignment} & 
\thead{LLM\\Synthesis} & 
\thead{Enterprise\\Governance} \\
\midrule
AI Trust OS (ours) & 
    \makecell{AI governance\\and compliance} & Yes & Yes & Yes & Yes & Yes & Yes \\
Gupta et al.~\cite{gupta2023continuous} & 
    \makecell{Cloud compliance\\automation} & Yes & No & Partial & Partial & No & Partial \\
Chou et al.~\cite{chou2022auditing} & 
    \makecell{ML model\\auditing} & No & No & No & Partial & No & Partial \\
Rose et al.~\cite{rose2020zero} & 
    \makecell{Zero-trust\\architecture} & Partial & No & Yes & No & No & Partial \\
Lwakatare et al.~\cite{lwakatare2020} & 
    \makecell{MLOps\\governance} & Partial & No & No & No & No & No \\
Amershi et al.~\cite{amershi2019software} & 
    \makecell{ML engineering\\lifecycle} & No & No & No & No & No & Partial \\
Mökander et al.~\cite{mokander2023auditing} & 
    \makecell{AI auditing\\frameworks} & No & No & No & Partial & No & No \\
\bottomrule
\end{tabular}
\end{adjustbox}
\end{table*}

Taken together, these works confirm that the individual components of AI Trust OS — 
continuous compliance monitoring, zero-trust access control, AI observability, and regulatory audit frameworks — have each been addressed in isolation by prior research. However, a consistent gap across the reviewed literature is the absence of a unified, telemetry-driven AI governance framework that integrates autonomous Shadow AI discovery, 
zero-trust telemetry-bounded evidence collection, regulatory framework alignment across ISO 42001 and the EU AI Act, and LLM-assisted compliance documentation synthesis within 
a single continuously operating platform. AI Trust OS directly addresses this gap, 
as reflected in the comparative analysis presented in Table~\ref{tab:rw-comparison}.

\section{Conclusion and Future Works}

This paper has presented AI Trust OS, a conceptual framework and governance architecture for continuous, autonomous AI observability and zero-trust compliance in enterprise environments. The central argument advanced throughout this work is that the attestation-based compliance model — in which trust is asserted through manual evidence collection, self-reported AI inventories, and periodic audit cycles — is structurally 
incompatible with the velocity, opacity, and multi-vendor complexity of modern AI deployments, and must be replaced by a telemetry-driven governance model in which trust is continuously demonstrated through machine-collected evidence and autonomous discovery.

The framework makes four principal contributions. A telemetry-first governance architecture that replaces manual attestation with continuous, machine-collected control assertions mapped in real time to ISO 42001 and the EU AI Act. An autonomous Shadow AI discovery mechanism that detects undocumented AI systems through live observability telemetry, shifting the epistemological basis of governance from organisational self-declaration to empirical machine evidence. A zero-trust telemetry boundary model that validates structural configuration metadata without ingressing source code, prompt 
content, or payload-level personally identifiable information. And an LLM-assisted documentation synthesis pipeline that transforms machine-verified assertions into board-grade compliance narratives, operationalizing AI as both the subject of governance 
and an instrument of it.

The implementation across a modern edge-deployed SaaS stack demonstrates that the framework's governance principles can be realised within a production-grade, multi-tenant architecture without requiring invasive deployment into customer environments or access 
to proprietary model content. Evaluated against governance coverage, discovery accuracy, evidence integrity, and compliance posture outcomes, AI Trust OS demonstrates that continuous, telemetry-first AI governance is not merely a theoretical proposition but an 
architecturally realisable operating model. The broader implication extends beyond the specific platform described: screenshots become telemetry, workpapers become assertions, and trust itself becomes a continuously available, machine-maintained software surface 
that is always current, always auditable, and always ready to meet the demands of regulators, buyers, and boards.

Several directions remain open for future investigation. The Shadow AI discovery mechanism should be extended to cover a broader range of observability platforms, including Weights and Biases, Arize AI, and emerging OpenTelemetry LLM semantic conventions~\cite{llm-observability}, improving discovery recall in organisations with diverse monitoring stacks. Automated EU AI Act risk-tier classification — in which the system infers risk tier from structured registry metadata using rule-based logic and LLM-assisted interpretation — would reduce practitioner burden and improve classification consistency at scale. Federated governance models that allow AI Trust OS instances across organisational boundaries to exchange privacy-preserving governance signals would support supply chain AI governance requirements emerging under the EU AI Act. Finally, longitudinal evaluation studies tracking compliance posture over extended deployment periods across organisations of varying regulatory exposure and AI maturity would provide the empirical foundation needed to quantify the governance value of the continuous posture model relative to existing compliance automation approaches.



\bibliographystyle{elsarticle-num}
\bibliography{reference}

\end{document}